\documentclass[letterpaper]{article} 
\usepackage{aaai24}  
\usepackage{times}  
\usepackage{helvet}  
\usepackage{courier}  
\usepackage[hyphens]{url}  
\usepackage{graphicx} 
\usepackage{natbib}  
\usepackage{caption} 
\usepackage{algorithm}
\usepackage{algorithmic}

\usepackage{newfloat}
\usepackage{listings}
\usepackage{graphicx}
\usepackage{latexsym}
\usepackage{times}
\usepackage{latexsym}
\usepackage{soul}
\usepackage{url}
\usepackage[utf8]{inputenc}
\usepackage{graphicx}
\usepackage{amsmath}
\usepackage{amsthm}
\usepackage{booktabs}
\usepackage{algorithm}
\usepackage{algorithmic}
\usepackage{multirow}
\usepackage{amssymb}
\usepackage{bbding}
\usepackage{mathrsfs}
\usepackage{longtable}
\usepackage{subcaption}
\usepackage[T1]{fontenc}
\usepackage{enumitem}
\usepackage{float}

\usepackage[utf8]{inputenc}
\usepackage{makecell}
\usepackage{microtype}
\usepackage{times}
\usepackage{latexsym}
\usepackage{soul}
\usepackage{url}

\usepackage{newfloat}
\usepackage{listings}
\DeclareCaptionStyle{ruled}{labelfont=normalfont,labelsep=colon,strut=off} 
\floatstyle{ruled}
\newfloat{listing}{tb}{lst}{}
\floatname{listing}{Listing}
\title{Labels Need Prompts Too: Mask Matching for Natural Language \\Understanding Tasks}
\author{
Bo Li$^{1,2}$, 
Wei Ye$^{1}$\thanks{{} {} Corresponding author.}, 
Quansen Wang$^{3}$, 
Wen Zhao$^{1}$, 
Shikun Zhang$^{1}$ \\}

\affiliations{
$^1$National Engineering Research Center 
  for Software Engineering, Peking University \\
  $^2$School of Software and Microelectronics, Peking University\\
  $^3$Boston University \\
  {deepblue.lb@stu.pku.edu.cn}, {wye@pku.edu.cn},\\ {quansenw@bu.edu}, \{{zhaowen, zhangsk\}@pku.edu.cn}
}

\usepackage{bibentry}

\begin{document}

\maketitle

\begin{abstract}
Textual label names (descriptions) are typically semantically rich in many natural language understanding (NLU) tasks. In this paper, we incorporate the prompting methodology, which is widely used to enrich model input, into the label side for the first time. Specifically, we propose a Mask Matching method, which equips an input with a prompt and its label with another, and then makes predictions by matching their mask representations. We evaluate our method extensively on 8 NLU tasks with 14 datasets. The experimental results show that Mask Matching significantly outperforms its counterparts of fine-tuning and conventional prompt-tuning, setting up state-of-the-art performances in several datasets. Mask Matching is particularly good at handling NLU tasks with large label counts and informative label names. As pioneering efforts that investigate the label-side prompt, we also discuss open issues for future study.
\end{abstract}

\section{Introduction}\label{sec:intro}

Large-scale pre-trained language models (PLMs) such as BERT \cite{DBLP:conf/naacl/DevlinCLT19} and RoBERTa \cite{DBLP:journals/corr/abs-1907-11692} have achieved impressive performances on a wide range of natural language understanding (NLU) tasks, e.g., topic classification \cite{DBLP:conf/naacl/XuLA22,DBLP:conf/acl/0002GLZH22}, sentiment analysis \cite{DBLP:conf/acl/ZhangZZZL0C22}, information extraction \cite{DBLP:conf/coling/LiYSXXZ20,DBLP:conf/acl/0001LDXLHSW22}, natural language inference \cite{DBLP:conf/acl/Dawkins21,DBLP:conf/acl/NighojkarL20}, and stance detection \cite{DBLP:conf/acl/LiuLTW21,DBLP:conf/sigir/JiangGSC22}. 

\begin{figure}[t]
	\centering
	\includegraphics[width=0.99\linewidth]{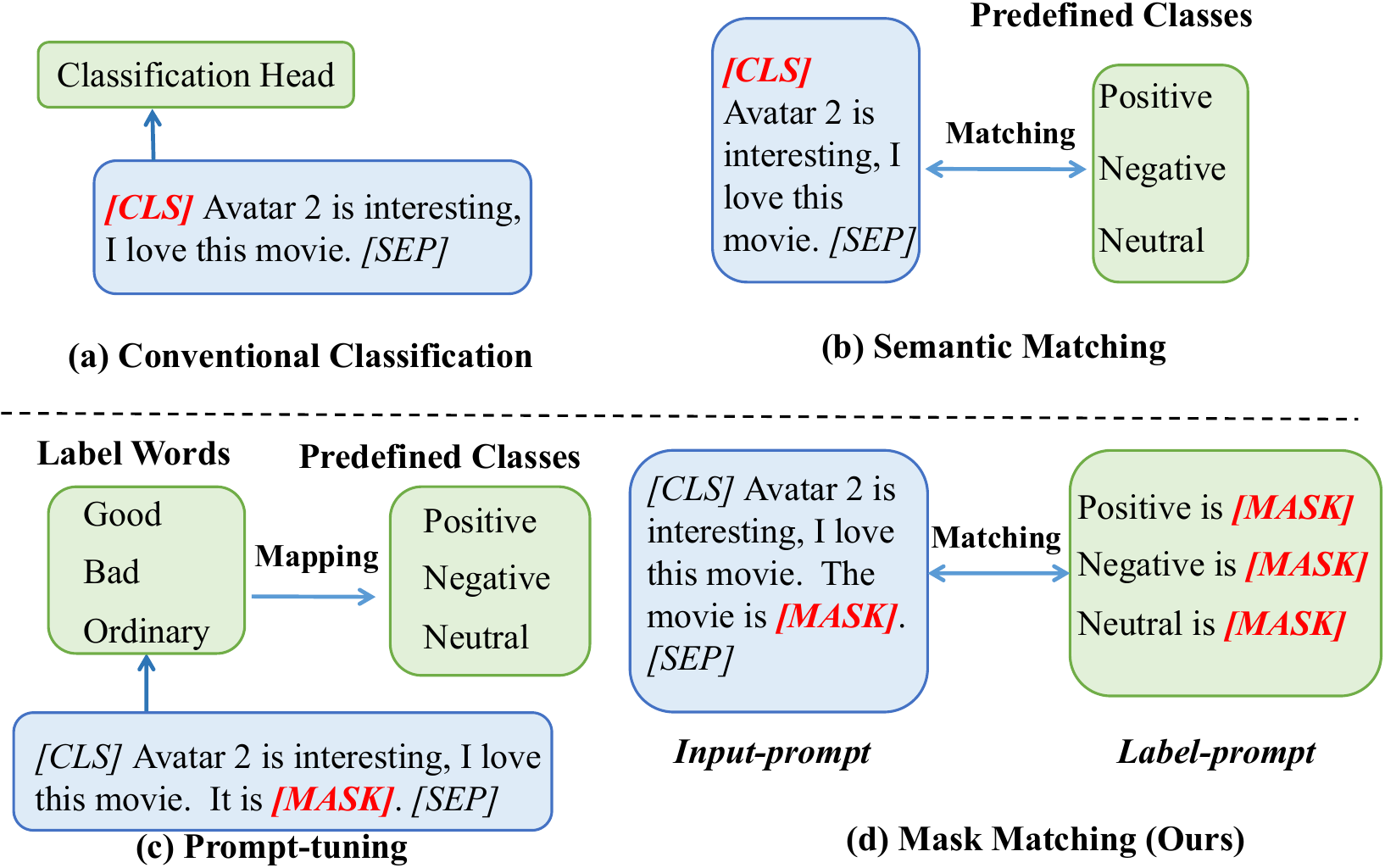}
	\caption{Conceptual illustration of our Mask Matching method and other popular paradigms. We use sentiment analysis as the example task.}
	\label{pic:intro}
\end{figure}

In general, fine-tuning PLMs with a classification head on the downstream dataset is the dominant solution for most NLU tasks, as shown in Figure \ref{pic:intro} (a). While this paradigm achieves impressive performances, it can not utilize textual semantics implied in label descriptions, which, however, have been proven to be beneficial for many downstream tasks \cite{DBLP:conf/naacl/XuLA22,DBLP:conf/coling/ArcoVK22,DBLP:conf/sigir/JiangGSC22,DBLP:conf/acl/LiangZL000X22,DBLP:conf/naacl/ObeidatFST19,DBLP:conf/naacl/HuangLXC22,DBLP:conf/emnlp/SainzLLBA21,DBLP:journals/corr/abs-2212-14266}. In this case, two other paradigms of NLU tasks can come to the rescue to some extent.

Semantic matching \cite{DBLP:conf/emnlp/SainzLLBA21,DBLP:conf/naacl/HuangLXC22,DBLP:conf/coling/ArcoVK22} involves generating representations of inputs and labels, and making predictions based on their semantic distances, as illustrated in Figure \ref{pic:intro} (b). This paradigm can naturally exploit semantic information from labels, e.g., utilizing representations of label names generated by PLMs. However, this paradigm heavily relies on label representations, and conventional modeling (e.g., with max-pooling or average-pooling) of label-related texts might not yield optimal ones.

Prompt-tuning \cite{DBLP:journals/corr/abs-2107-13586,DBLP:journals/corr/abs-2105-11259,DBLP:conf/www/ChenZXDYTHSC22} naturally exploits textual semantics of labels by manually designing proper label verbalizer \cite{DBLP:conf/eacl/SchickS21,DBLP:conf/acl/GaoFC20,DBLP:conf/acl/LeeKTAF0MSPR22}, as shown in Figure \ref{pic:intro} (c). Nevertheless, selecting label words is non-trivial and labor-intensive for many tasks, such as entity typing \cite{DBLP:conf/acl/DingXCWHXZL20} and relation extraction \cite{DBLP:conf/emnlp/ZhangZCAM17}. Some researchers explore utilizing trainable virtual label words instead \cite{DBLP:journals/corr/abs-2204-13413,DBLP:conf/www/ChenZXDYTHSC22,DBLP:journals/corr/abs-2105-11259,DBLP:conf/acl/ParkJKKN22}. Though avoiding cumbersome verbalizer engineering, they completely discard label-related text information.

In this research, we present a new paradigm called Mask Matching that mines label semantic information using the prompting methodology on the label side. To capture the semantics of inputs and labels, we introduce a \textit{label-prompt} in addition to the \textit{input-prompt}, which results in effective predictions based on a simple matching strategy. In this way, the semantics of inputs and labels are captured by the mask representations from their corresponding prompts. Mask Matching combines the merits of traditional semantic matching and prompt-tuning paradigms and avoids the verbalizer engineering, yielding a conceptually simple and easy-to-implement method.

To verify the effectiveness of Mask Matching, we conduct extensive experiments on 8 NLU tasks across 14 datasets. Our method demonstrated remarkable performances on both full training setting (\S \ref{sec:full}) and low-resource settings (\S \ref{sec:low_resource}) compared to its counterparts of fine-tuning and prompt-tuning. Additionally, it achieves comparable or better results than many state-of-the-art methods. Notably, Mask Matching exhibits more evident superiority when the predefined class in the datasets is numerous, and their names are informative. Besides the performance improvements, we also discuss the potential research directions upon Mask Matching (\S \ref{sec:discussion}). Below we summarize our main contributions:

\begin{itemize}
    \item We propose Mask Matching, a new natural language understanding (NLU) paradigm that simultaneously performs prompting on inputs and labels. It can be easily and effectively applied in most NLU tasks by matching the two mask representations of both sides.

    \item Extensive experiments show that Mask Matching significantly outperforms its counterparts of fine-tuning and prompt-tuning, and achieves competitive results compared with recent state-of-the-art models. 

    \item As pioneering efforts that investigate the label-side prompt, we propose many open problems to inspire future studies in this direction.
\end{itemize}

\section{Related Work}

As this work aims to explore utilizing semantic information in label names, we briefly introduce how existing approaches utilize label semantic information.

Label semantics is beneficial to many NLU tasks, such as text classification \cite{DBLP:conf/naacl/XuLA22,DBLP:conf/coling/ArcoVK22}, stance detection \cite{DBLP:conf/sigir/JiangGSC22,DBLP:conf/acl/LiangZL000X22}, named entity recognition \cite{DBLP:conf/naacl/ObeidatFST19,DBLP:conf/naacl/HuangLXC22} and relation classification \cite{DBLP:conf/emnlp/SainzLLBA21,DBLP:journals/corr/abs-2212-14266}. In recent works, researchers proposed several approaches to fully use the available semantic information in labels and achieve desirable performances, e.g., semantic matching methods and prompt-tuning methods.

The semantic matching method is the default solution for sentence-pair tasks, such as natural language inference \cite{DBLP:conf/emnlp/RajpurkarZLL16} and paraphrase \cite{DBLP:conf/semeval/XuCD15}. This method could utilize the label semantic via encoding label descriptions and achieves good results on tasks where labels contain rich semantic information \cite{DBLP:journals/corr/abs-2104-14690,DBLP:conf/emnlp/SainzLLBA21,DBLP:conf/naacl/HuangLXC22,DBLP:conf/coling/ArcoVK22}. Typically, Semantic matching method usually jointly encodes premise/hypothesis or input/labels and evaluates the relationship between both ends. This paradigm heavily relies on label representations, while conventional modeling (e.g., with max-pooling or average-pooling) of label-related texts might not yield optimal ones.

Prompt-tuning is an emerging paradigm in recent years. It could bridge the gap between pre-training and fine-tuning, showing surprising power on a wide range of NLP tasks \cite{DBLP:journals/corr/abs-2107-13586,DBLP:journals/corr/abs-2105-11259,DBLP:conf/www/ChenZXDYTHSC22}. Specially, with a carefully designed template, prompt-tuning transforms the target task to a cloze style format and outputs the prediction via a special mask token. Prompt-tuning naturally utilizes label semantic information by manually designing properly label verbalizer \cite{DBLP:conf/eacl/SchickS21,DBLP:conf/acl/GaoFC20,DBLP:conf/acl/LeeKTAF0MSPR22}, such as “good” for "positive" and “bad” for "negative". To avoid human involvement, some researchers also explore utilizing trainable virtual label words \cite{DBLP:journals/corr/abs-2204-13413,DBLP:conf/www/ChenZXDYTHSC22,DBLP:journals/corr/abs-2105-11259,DBLP:conf/acl/ParkJKKN22}. Although most prompt-tuning methods perform well in low-resource scenarios \cite{DBLP:conf/acl/LiuLTW21,DBLP:conf/acl/0003XSHTGJ22,DBLP:conf/ijcai/LiuCX22}, they still struggles to achieve on pair results compared with fine-tuning, especially when PLMs are relatively small and the training data is sufficient\cite{DBLP:conf/emnlp/LesterAC21,DBLP:conf/acl/GaoFC20,DBLP:journals/corr/abs-2208-03229}.

\section{Approach}

In this section, we first briefly introduce the preliminaries and the prompt-tuning method, then present our Mask Matching in detail.

\subsection{Prompt-tuning}\label{sec:PT}

Unlike traditional fine-tuning methods that utilize the \textit{[CLS]} token for NLU tasks \cite{DBLP:conf/naacl/DevlinCLT19,DBLP:journals/corr/abs-1907-11692}, prompt-tuning methods use a special mask token and a pre-defined template for prediction output. For instance, when dealing with a sentiment analysis task with input \textit{[$X$]}, researchers may use the following template: \textit{"[X]. It is [MASK]"}. Where the mask representation $M_I$ is then converted to a class prediction by a predefined label verbalizer, e.g., \{"good" → "positive", "bad" → "negative" and "ordinary" → "natural" \}. In the above example, "good", "bad" and "ordinary" are label words, and "positive", "negative" and "natural" are label names. An example of such a system can be seen in Figure \ref{pic:intro} (c). With a well-designed template and label verbalizer, prompt-tuning is effective in solving single-input tasks such as sentiment analysis and topic classification. 

Despite the above tasks, prompt-tuning could also handle paired-input tasks by concatenating two inputs with a prompt. We take the paraphrase task as an example. Given two sentences \textit{[$X_1$]} and \textit{[$X_2$]}, the prompt-tuning method will recompose inputs as \textit{[$\hat{X}$]}: \textit{[$X_1$] [$X_2$] The relation between two sentences is [MASK]}, 

Depending on the task, we can employ either manually chosen real words \cite{DBLP:conf/eacl/SchickS21,DBLP:conf/acl/GaoFC20,DBLP:journals/corr/abs-2105-11259,DBLP:conf/acl/LeeKTAF0MSPR22} or virtual label tokens that can be trained \cite{DBLP:journals/corr/abs-2204-13413,DBLP:conf/www/ChenZXDYTHSC22,DBLP:journals/corr/abs-2105-11259,DBLP:conf/acl/ParkJKKN22} as our label verbalizer. In this study, we opt for the latter, virtual token approach because it does not entail manual effort and obtains more consistent and better performance, particularly when the quantity of training data is substantial. To clarify, we refer to the aforementioned prompts as \textit{input-prompts} since they all exist on the input side.

 \begin{figure*}[t]
	\centering
	\begin{subfigure}{0.49\linewidth}
		\centering
		\includegraphics[width=0.99\linewidth]{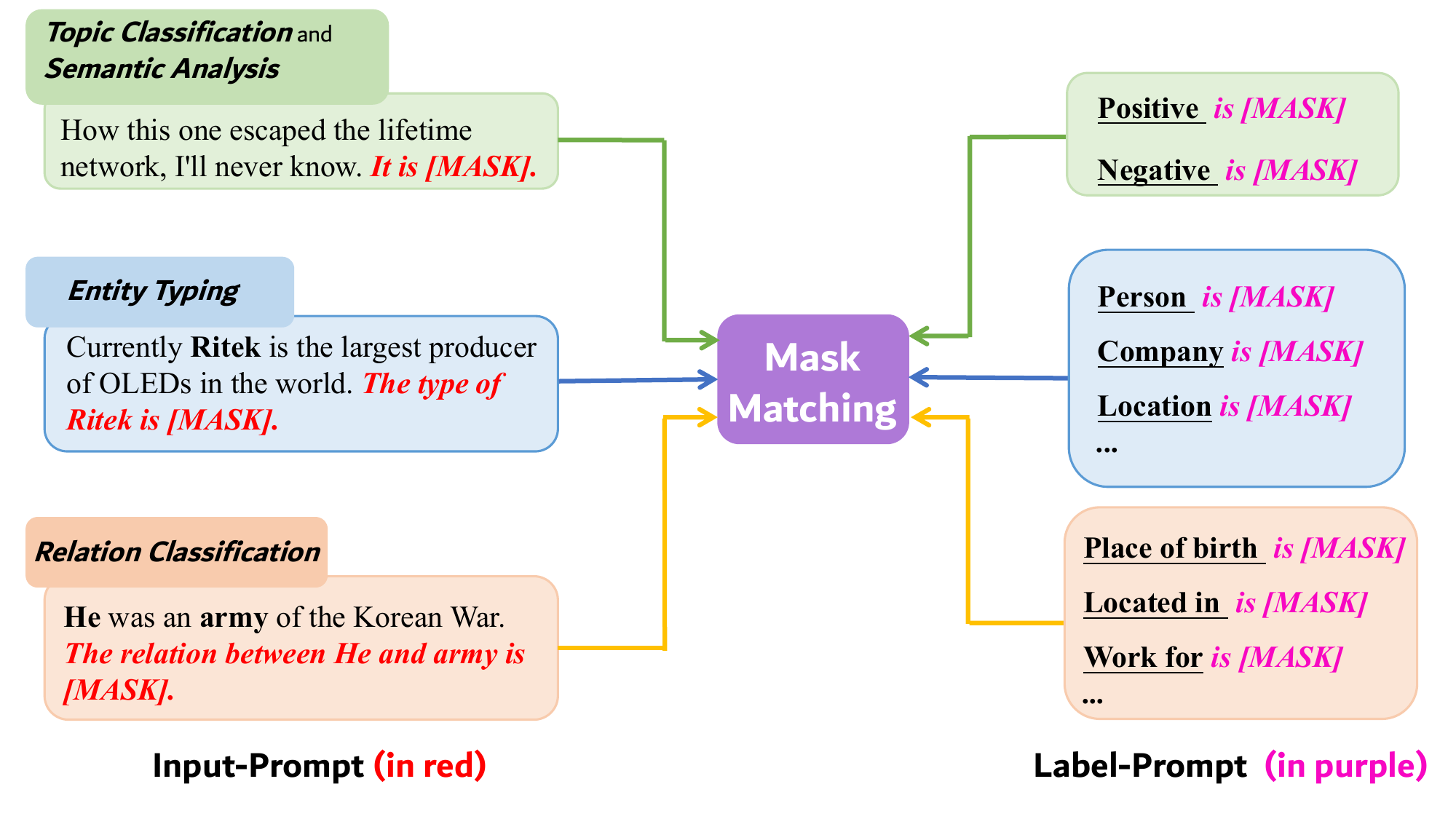}
		\caption{Single-Input Tasks}
		\label{chutian3}
	\end{subfigure}
	\centering
	\begin{subfigure}{0.49\linewidth}
		\centering
		\includegraphics[width=0.99\linewidth]{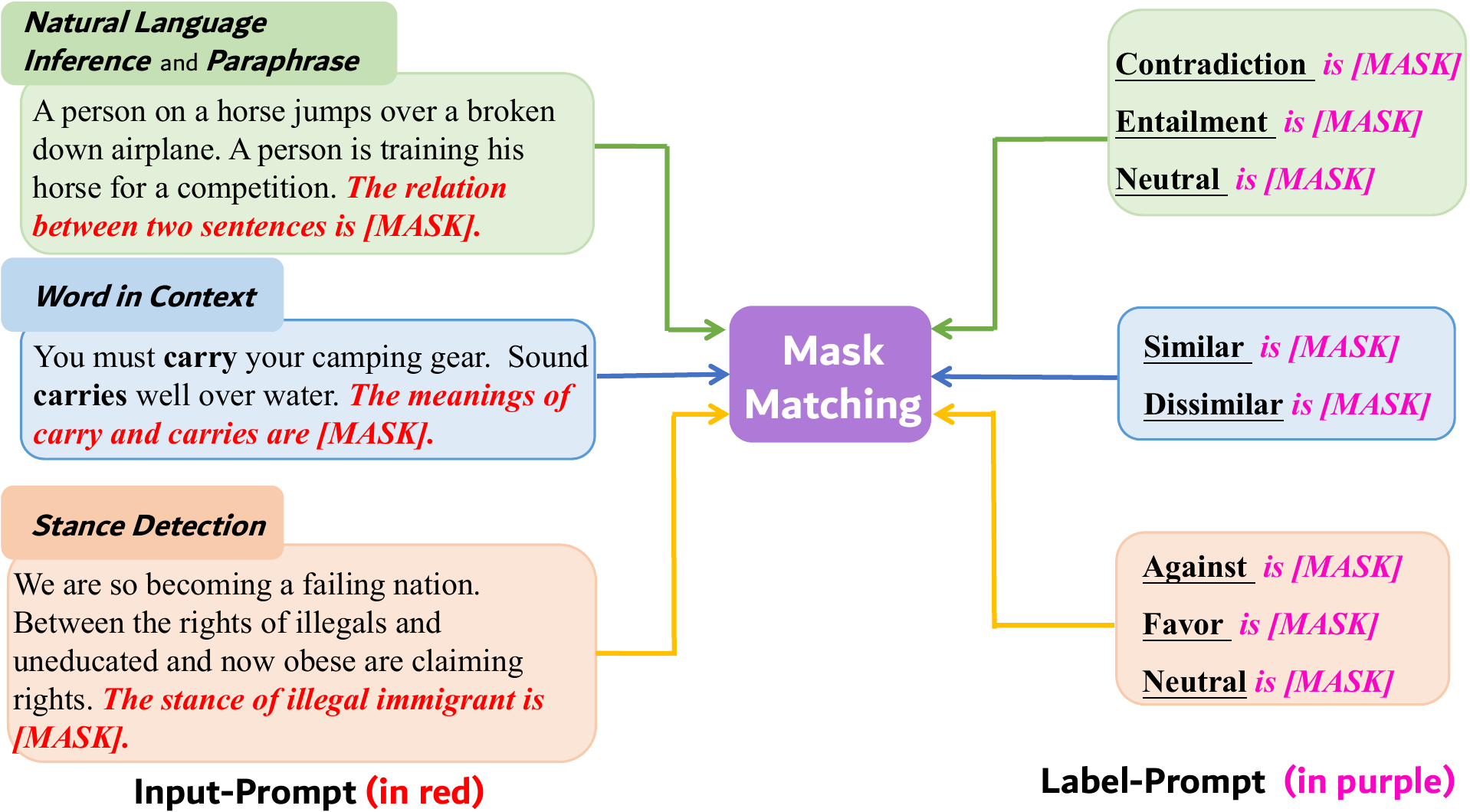}
		\caption{Paired-Input Tasks}
		\label{chutian3}
	\end{subfigure}
	\caption{These examples provide an intuitive understanding of several single-input and paired-input tasks, all of which are extracted from real-world datasets. The highlighted \textbf{bold} target entities, keywords, and phrases to be identified, and the corresponding labels are underlined. The \textit{input-prompts} are distinguished in red, while the \textit{label-prompts} are showcased in purple. The mask in each \textit{label-prompt} symbolizes the corresponding label. The cross-entropy loss in Mask Matching is calculated between the mask in the \textit{input-prompt} and all masks in the \textit{label-prompts}. It’s important to note that the entire PLM is trainable. Better viewed in color.}
	\label{pic:model}
\end{figure*}

\subsection{Mask Matching}\label{sec:mask_match}

This section outlines Mask Matching, a new method that utilizes two mask tokens to learn more useful information from both the input and label sides. Overall, in addition to the \textit{input-prompt} $P_I$ mentioned in \S \ref{sec:PT}, Mask Matching involves a \textit{label-prompt} $P_L$. For each label, \textit{label-prompt} adds a template with a mask token after the label name. The mask representation $M_L$ from $P_L$ is used as the label representation. At the training phase, Mask Matching optimizes parameters by computing the cross-entropy loss between $M_I$ and $M_L$ over all predefined labels. Note that we use the same PLM to encode inputs and labels.


The \textit{label-prompt} in Mask Matching eliminates the need for label verbalizer construction and enhances the utilization of semantic information in label names. The subsequent section explains how to employ Mask Matching for various NLU tasks. Based on the input format, we divide the NLU tasks examined in this research into two categories: \textbf{Single-input Tasks} and \textbf{Paired-input Tasks}.

\textbf{Single-input Tasks.} This type contains a wide range of NLU tasks with a single input, such as topic classification, sentiment analysis, entity typing, and relation classification. In these tasks, Mask Matching incorporates prompts on both the input and label sides simultaneously. As shown in Figure \ref{pic:model} (a), we use slightly different \textit{input-prompt} variations for different tasks, and the \textit{label-prompt} $P_L$ is consistent across all tasks, defined as $P_L$ = \textit{is [MASK].}

\begin{itemize}
    \item \textbf{Topic Classification and Sentiment Analysis.} In these tasks, we need to classify which class the input text [$X$] belongs to. We set the \textit{input-prompt} $P_I$ = \textit{It is [MASK]}, as shown in Figure \ref{pic:model} (a).
    
    \item \textbf{Entity Typing.} This task requires identifying the entity type for a given target entity. As shown in Figure \ref{pic:model} (a), the input text \textit{[X] = Currently Ritek is the largest producer of OLEDs in the world}, and the target entity is \textit{Ritek}. We use the following simple template in the \textit{input-prompt}, where $P_I$ = \textit{The type of [target entity] is [MASK].}

    \item \textbf{Relation Classification.} This task aims to classify the relation between \textit{head entity} and \textit{tail entity} in a given input. As shown in Figure \ref{pic:model} (a), the input text \textit{[X] = He was an army of the Korean War}, and the \textit{head entity} and \textit{tail entity} are \textit{He} and \textit{army}. The \textit{input-prompt} $P_I$ = \textit{The relation between [head entity] and [tail entity] is [MASK]}. Note that we also add entity markers and entity type information to the input, as previous works show these two types of information could bring huge improvements \cite{DBLP:conf/acl/SoaresFLK19,DBLP:conf/acl/TianCSW20}.
\end{itemize}

\textbf{Paired-input Tasks.} These tasks need to identify the relationship between two given text pieces, e.g., neutral language inference, paraphrase, word in context and stance detection. For the input side, we concatenate two inputs with a task-specific \textit{input-prompt} $P_I$. As for the label side, we use the same \textit{label-prompt} $P_L$ as described in Single-input Tasks.


\begin{itemize}
    \item \textbf{Natural Language Inference and Paraphrase.} The above two tasks aim to distinguish the relationship between two sentences. We first concatenate two given sentences \textit{[$X_1$]} and \textit{[$X_2$]}, then add a simple \textit{input-prompt} $P_I$, where $P_I$ = \textit{The relation between two sentences is [MASK].} An example is shown in Figure \ref{pic:model} (b).
    
    \item \textbf{Word in Context.} This is a semantic distinction task with paired-input. In this task, we have two input texts and two keywords, \textit{[$K_1$]} and \textit{[$K_2$]}, often the same word, but in different tenses or morphemes. The objective is to determine if the two keywords have similar meanings. For example, given the inputs \textit{[$X_1$] = You must carry your camping gear} and \textit{[$X_2$] = Sound carries well over water}, with \textit{carry} and \textit{carries} as the two keywords, the \textit{input-prompt} $P_I$ is added to the concatenated input, where $P_I$ = \textit{[$K_1$] is [MASK] to [$K_2$]}. This task is demonstrated in Figure \ref{pic:model} (b).

    \item \textbf{Stance Detection.} Here, we are asked to identify whether a text is in favor of, against, or neutral to a given target (e.g., an event, or a claim). As shown in Figure \ref{pic:model} (b), the given input text is \textit{[$X$] = We are so becoming a failing nation. Between the rights of illegals and uneducated and now obese are claiming rights.} The target phrase is \textit{[T]} = \textit{illegal immigrant}. For the given input \textit{[X]} and the target \textit{[T]}, we add a \textit{input-prompt} $P_I$ after \textit{[X]}, where \textit{$P_I$ = {The stance of \textit{[T]} is [MASK].}}
    
\end{itemize}

Note that the label names of these tasks are all meaningful words or phrases. Thus we can directly use the \textit{label-prompt} to learn useful semantic information from label names.

\subsection{Training and Testing}

During the training phase, for a given sample with $m$ predefined classes, we utilize the mask within the \textit{input-prompt} as the input representation and $m$ corresponding \textit{[MASK]s} from the \textit{label-prompts} as the label embeddings. We compute the cross-entropy loss between the input representation and label embeddings to compute the loss and optimize the entire PLM. For instance, let's consider the entity typing task, as illustrated in Figure 2(a). Assuming that the \textit{[MASK]} representation from the input prompt is denoted as $h$, and the \textit{[MASK]} representations from the label prompt are denoted as $m_{1}$, $m_{2}$, ..., $m_{n}$, where $n$ represents the number of predefined labels. To begin, we derive the prediction probability using a softmax activation function as follows:

\begin{equation}
    p = softmax(h \cdot [m_{1}, m_{2}, ..., m_{n}]^T),
\end{equation}
where [*] means concatenation.
Subsequently, we optimize the \textit{cross-entropy (CE) loss} between \textit{p} and the ground-truth label \textit{q} (one-hot format):
\begin{equation}
    loss = CE(p, q),
\end{equation} 

During testing, we calculate the dot product between the input representation and all label embeddings to generate the final prediction.
\section{Experimental Setup}

\subsection{Datasets}
We use 8 different natural language understanding tasks across 14 datasets to verify the effectiveness of our proposed method. The metric and class numbers for each dataset are shown in Table \ref{table:data}. These datasets are chosen from a wide range of common NLU tasks. Some of them are selected from GLUE benchmark \cite{DBLP:conf/iclr/WangSMHLB19} and SuperGLUE \cite{DBLP:conf/nips/WangPNSMHLB19}, others are popular in various specific research fields, such as entity typing, relation classification, and stance detection. The first category is \textit{single-input task}, such as topic classification (R8 and R52\footnote{https://www.cs.umb.edu/~smimarog/textmining/datasets/}, we use the data split proposed by \cite{DBLP:journals/corr/abs-2105-05727}), sentiment analysis (MR \cite{DBLP:conf/acl/PangL05} and IMDb \cite{DBLP:conf/acl/MaasDPHNP11}), entity typing (FEW-NERD \cite{DBLP:conf/acl/DingXCWHXZL20} and BBN \cite{weischedel2005bbn,DBLP:conf/kdd/Huang0022}) and relation classification (TACRED \cite{DBLP:conf/emnlp/ZhangZCAM17} and TACRED-Revisited \cite{DBLP:conf/acl/AltGH20a}). We also explore applying Mask Matching to several \textit{paired-input tasks}, including natural language inference (QNLI \cite{DBLP:conf/emnlp/RajpurkarZLL16} and SNLI \footnote{https://nlp.stanford.edu/projects/snli/} \cite{DBLP:conf/emnlp/BowmanAPM15}), paraphrase (PIT2015 \cite{DBLP:conf/semeval/XuCD15} and QQP \footnote{https://www.quora.com/q/quoradata/}), word in context (WiC \cite{DBLP:conf/naacl/PilehvarC19}), and stance detection (VAST \cite{allaway2020zero}). For a fair comparison, all the datasets and the data split are the same as in previous works.

\subsection{Comparison Methods}

In this research, we compare our method with the following approaches:

\textbf{State-of-the-art} method on the single dataset. We report the previous 
best results among all the datasets, and these results are directly cited from public papers. It is important to note that our comparisons primarily focus on models that utilize widely used pre-trained language models like BERT, RoBERTa, and LUKE. However, some state-of-the-art methods that rely on pre-training with domain-specific datasets are not directly comparable to our approach. 

\textbf{Fine-tuning} PLMs on each dataset. This is the classical solution for NLU tasks. We use the classification-based method for \textit{single-input} tasks, and the semantic matching method for \textit{paired-input} tasks. 

\textbf{Prompt-tuning} utilizes a trainable virtual embedding for each label in a given task while keeping other settings the same as Fine-tuning. The primary reason for using a trainable virtual label embedding instead of manually selecting label words is that it requires no manual effort, making it more generalizable to various tasks. Additionally, in many tasks that have a large number of predefined classes, it is often challenging to choose a single discriminating word for each label, particularly in entity typing and relation classification (\S \ref{sec:intro}).

\textbf{Semantic Matching} has similar training and testing procedures with Mask Matching. The only difference is that Semantic Matching uses the \textit{max-pooling} over label name as the label representation. \footnote{We also explore using the \textit{[CLS]} token in the input side or the \textit{average-pooling} as alternates, but we found that \textit{max-pooling} is the best choice.} The label names are the same as we used in \textbf{Mask Matching}.

\begin{table}
\caption{Here is a summary of the datasets we evaluated in our research, with \#C representing the number of classes in each dataset. We tested our models on 8 different tasks across 14 datasets, and our metric choices align with those used in previous studies.}
\label{table:data}
\renewcommand{\arraystretch}{1.2}
\centering
\setlength{\tabcolsep}{1.5mm}
\begin{tabular}{c|c|c|c}
\toprule[1.5pt]
\textbf{Task}                                                                                  & \textbf{Metric}                          & \textbf{Dataset} & \textbf{\#C} \\ \hline
\multirow{2}{*}{\textbf{\begin{tabular}[c]{@{}c@{}}Topic\\ Classification\end{tabular}}}       & \multirow{2}{*}{\textit{Accuracy}}       & R8               & 8              \\
                                                                                               &                                          & R52              & 52             \\\hline
\multirow{2}{*}{\textbf{\begin{tabular}[c]{@{}c@{}}Semantic\\ Analysis\end{tabular}}}          & \multirow{2}{*}{\textit{Accuracy}}       & MR               & 2                \\
                                                                                               &                                          & IMDb             & 2                \\\hline
\multirow{2}{*}{\textbf{Entity Typing}}              & \multirow{2}{*}{\textit{Loose Micro-F1}} & FEW-NERD         & 66             \\
                                                                                               &                                          & BBN              & 47             \\\hline
\multirow{2}{*}{\textbf{\begin{tabular}[c]{@{}c@{}}Relation\\ Classification\end{tabular}}}    & \multirow{2}{*}{\textit{Micro-F1}}       & TACRED           & 42             \\
                                                                                               &                                          & TAC-REV          & 42             \\ \bottomrule[1.5pt]
\multirow{2}{*}{\textbf{\begin{tabular}[c]{@{}c@{}}Natural Language\\ Inference\end{tabular}}} & \multirow{2}{*}{\textit{Accuracy}}       & QNLI             & 2              \\
                                                                                               &                                          & SNLI             & 3              \\\hline
\multirow{2}{*}{\textbf{Paraphrase}}                                                           & \multirow{2}{*}{\textit{Micro-F1}}       & PIT2015          & 2              \\
                                                                                               &                                          & QQP              & 2              \\\hline
\textbf{Word in Context}                                                                       & \textit{Accuracy}                        & WiC              & 2                \\\hline
\textbf{Stance Detection}                                                                      & \textit{Macro-F1}                        & VAST             & 3\\\bottomrule[1.5pt]             
\end{tabular}
\end{table}

\begin{table*}[t]
\caption{Performance of different methods on 14 NLU datasets under the full training setting. We re-implemented some of the methods and cited comparable SOTA results from public papers. The best performances achieved by Mask Matching are denoted in \textbf{bold}. Since previous methods in PIT2015 and VAST datasets used specific PLMs such as SBERT \cite{DBLP:conf/emnlp/ReimersG19}, we left the comparable SOTA results blank. To ensure a fair comparison, we report the performance on the development sets of WiC, QNLI, SNLI, and QQP, as previous works \cite{DBLP:journals/corr/abs-1907-11692,DBLP:journals/corr/abs-2111-09543,DBLP:journals/corr/abs-2104-14690,DBLP:journals/corr/abs-2204-06644} only presented the single model’s performances on the development set. The numbers in parentheses indicate the improvement in performance compared with Fine-tuning. We averaged the results of our methods over three random seeds, and the results we reported are statistically significant with $p < 0.05$.}
\label{table:main}
\renewcommand{\arraystretch}{1.2}
\centering
\setlength{\tabcolsep}{2mm}
\begin{tabular}{c|c|c|c|c|c|c}
\toprule[1.5pt]
\textbf{Task}                                                                                                  & \textbf{Dataset} & \textbf{Comparable SOTA}     & \textbf{\begin{tabular}[c]{@{}c@{}}Fine-\\ tuning\end{tabular}} & \textbf{\begin{tabular}[c]{@{}c@{}}Prompt-\\ tuning\end{tabular}} & \textbf{\begin{tabular}[c]{@{}c@{}}Semantic \\ Matching\end{tabular}} & \textbf{\begin{tabular}[c]{@{}c@{}}Mask \\ Matching\end{tabular}} \\ \hline
\multirow{2}{*}{\textit{\textbf{\begin{tabular}[c]{@{}c@{}}Topic\\ Classification\end{tabular}}}}           & R8                   & 98.2 \cite{DBLP:journals/corr/abs-2105-05727}                     & 98.1                                                            & 98.0                                                              & 98.0                                                           & \textbf{98.3(+0.2)}                                            \\
                                                                                                                & R52                & 96.6 \cite{DBLP:journals/corr/abs-2105-05727}                     & 96.4                                                            & 96.7                                                              & 96.5                                                          & \textbf{96.9(+0.5)}                                            \\ \hline
\multirow{2}{*}{\textit{\textbf{\begin{tabular}[c]{@{}c@{}}Sentiment\\ Analysis\end{tabular}}}}      & MR                     & 92.5 \cite{DBLP:journals/corr/abs-2104-14690}                      & 91.9                                                            & 91.9                                                              & 92.0                                                           & 92.3(+0.4)                                                     \\
                                                                                                              & IMDb                 & 97.1 \cite{DBLP:conf/acl/DingSWSTW020}                     & 96.4                                                            & 96.4                                                              & 96.4                                                           & 96.6(+0.2)                                                     \\ \hline                                                                                                                
                                                                                                                
\multirow{2}{*}{\textit{\textbf{\begin{tabular}[c]{@{}c@{}}Relation \\ Classification\end{tabular}}}}        & TACRED                    & 75.6 \cite{DBLP:conf/aaai/Li0ZZ23}                     & 74.4                                                            & 74.3                                                              & 73.2                                                           & 75.2(+0.8)                                            \\
                                                                                                             & TAC-REV                      & 84.1 \cite{DBLP:conf/aaai/Li0ZZ23}                    & 83.2                                                            & 83.3                                                              & 82.1                                                           & \textbf{84.4(+1.2)}                                            \\ \hline
\multirow{2}{*}{\textit{\textbf{Entity Typing}}}               & Few-NERD            & 85.7  \cite{DBLP:journals/corr/abs-2108-10604}                    & 84.6                                                            & 84.8                                                              & 84.4                                                           & 85.2(+0.6)                                                     \\
                                                                                                              & BBN                  & 82.2 \cite{DBLP:conf/kdd/Huang0022}                     & 80.3                                                            & 80.0                                                              & 79.4                                                           & 81.2(+0.9)                                                     \\ \hline
\multirow{2}{*}{\textit{\textbf{\begin{tabular}[c]{@{}c@{}}Natural Language\\ Inference\end{tabular}}}}               & QNLI(dev)                & 96.5  \cite{DBLP:journals/corr/abs-2204-06644}                    & 94.6                                                            & 94.5                                                              & 94.2                                                           & 94.9(+0.3)                                                     \\
                                                                                                               & SNLI(dev)                  & 93.1 \cite{DBLP:journals/corr/abs-2104-14690}& 92.0                                           & 92.0                                             & 91.7                                          & 92.3(+0.3)                                           \\ \hline
\multirow{2}{*}{\textit{\textbf{Paraphrase}}}                                                                      & PIT2015               & -                         & 83.6                                                            & 83.7                                                              & 83.7                                                           & \textbf{84.1(+0.5)}                                            \\
                                                                                                             & QQP(dev)                 & 93.2 \cite{DBLP:journals/corr/abs-2204-06644}                     & 92.2                                                            & 92.1                                                              & 91.8                                                           & 92.4(+0.2)                                                     \\ \hline
\textit{\textbf{Word in Context}}                                                                                    & WiC(dev)               & 71.1  \cite{DBLP:conf/conll/Liu0CKV21}                    & 67.6                                                            & 68.3                                                              & 66.1                                                           & 69.3(+1.7)                                                     \\ \hline                                                                                                             
\textit{\textbf{Stance Detection}}                                                                                  & VAST                   & -                         & 76.8                                                            & 77.3                                                              & 77.1                                                           & \textbf{78.0(+1.2)}                                            \\ \bottomrule[1.5pt]
\end{tabular}
\end{table*}

\subsection{Experimental Setup}
We use Pytorch \cite{DBLP:conf/nips/PaszkeGMLBCKLGA19} and Tesla T4 GPU in our experiments. To ensure the simplicity of our framework, we maintain consistent hyper-parameters across all experiments and observe that results from Mask Matching are consistent across various settings. Specifically, we implement a batch size of 8, with a gradient accumulation of 4, and employ the AdamW optimizer \cite{DBLP:conf/iclr/LoshchilovH19}, with a learning rate of 1e-5 and a warm-up ratio of 0.2 \cite{DBLP:journals/corr/GoyalDGNWKTJH17} for all datasets. We use RoBERTa-large \footnote{https://huggingface.co/roberta-large} in all tasks, except for the entity typing task, where previous findings demonstrated that BERT-large \footnote{https://huggingface.co/bert-lagre-uncased} offered superior performance. The training epoch is set to 20, and the maximum input length is limited to 500. While we consider several alternative templates for the \textit{input-prompt} and \textit{label-prompt} in addition to the default template described in \S \ref{sec:mask_match}, we observe that the performance differences were negligible (\S \ref{sec:diff_prompt}). To reduce variability, we run each model three times under full training setting and five times in the low-resource scenario. We utilize identical experimental settings for Fine-tuning, Prompt-tuning, Semantic Matching, and Mask Matching.

\section{Results}\label{sec:results}

\subsection{Full Training Setting}\label{sec:full}


The results of the full training setting on 8 NLU tasks across 14 datasets are presented in Table \ref{table:main}, with the improvements of Mask Matching over Fine-tuning shown in parentheses. Our key observations are listed below:

First, the effectiveness of Mask Matching in comparison to Fine-tuning and Prompt-tuning is remarkable. The performance improvement ranges from 0.2\% to almost 2\%. For datasets that include label names containing rich semantic information, the improvements are more significant. For example, Mask Matching improves the F1 score from 83.2\% to 84.4\% on TAC-REV, and from 80.0\% to 81.2\% on BBN. But the improvements on R8, IMDb, QQP, and natural language inference tasks are relatively small. We attribute this to the following reasons: 1) Strong baseline performances have limited room for improvement; 2) Current design for Mask Matching does not cope well with sentence-paired tasks.\footnote{In fact, this is a common problem faced by prompt-tuning-based methods \cite{DBLP:conf/nips/BrownMRSKDNSSAA20,DBLP:journals/corr/abs-2103-10385,DBLP:conf/eacl/SchickS21,DBLP:conf/acl/TabasiRP22}.} Besides, Mask Matching also outperforms Prompt-tuning significantly by fully utilizing the label semantic information. The above results show the effectiveness of Mask Matching, and Mask Matching could serve as a strong baseline for a wide range of NLU tasks.

Second, our intention is not to establish a new state-of-the-art; nevertheless, we report the current SOTA for individual datasets to exhibit the difference between our method and the current leading models. As far as we know, most of these SOTA methods involve task-specific components or training strategies. However, Mask Matching still achieves competitive performances and even obtains slightly better results than comparable SOTA in several tasks, such as topic classification and relation classification. We find that these tasks have a relatively large number of labels, and their labels contain rich information, which is consistent with our conclusion in the previous analysis.


\begin{table}[h]
\caption{Performance of different methods on 14 NLU datasets under the low-resource setting. We randomly select 10\% percent of the whole training set, and keep the development and test sets unchanged. Results are averaged over five random seeds to reduce the randomness.}
\label{tabel:low}
\centering
\setlength{\tabcolsep}{3mm}
\begin{tabular}{c|c|c}
\toprule[1.5pt]
\textbf{Dataset}  & \textbf{Prompt-tuning} & \textbf{Mask Matching} \\ \hline
\textbf{R8}       & 97.3                   & 97.4(+0.1)             \\
\textbf{R52}      & 92.5                   & 93.0(+0.5)             \\
\textbf{MR}       & 95.7                   & 95.9(+0.2)             \\
\textbf{IMDb}     & 90.8                   & 91.4(+0.6)             \\
\textbf{FEW-NERD} & 74.7                   & 82.2(+7.5)             \\
\textbf{BBN}      & 74.3                   & 79.6(+5.3)             \\
\textbf{TACRED}   & 64.9                   & 67.0(+2.1)             \\
\textbf{TAC-REV}  & 72.1                   & 73.2(+1.1)             \\ \hline
\textbf{QNLI}     & 91.7                   & 92.1(+0.4)             \\
\textbf{SNLI}     & 90.2                   & 90.9(+0.7)             \\
\textbf{PIT2015}  & 75.4                   & 78.3(+2.9)             \\
\textbf{QQP}      & 88.7                   & 88.7(+0.0)             \\
\textbf{WiC}      & 58.8                   & 62.4(+3.6)             \\
\textbf{VAST}     & 75.5                   & 73.8(-1.7)             \\\bottomrule[1.5pt] 
\end{tabular}
\end{table}

Finally, Semantic Matching is another critical baseline with similar training and testing procedures to Mask Matching. Our experimental results demonstrate that Semantic Matching provides a weaker and less accurate label representation than Mask Matching, despite exploiting the same label semantic information. In particular, Semantic Matching performs poorly on relation classification and entity typing tasks. We attribute this to the impact of word frequency on the representation~\cite{DBLP:journals/corr/abs-2201-04337,DBLP:conf/acl/ZhouECJ22,DBLP:journals/corr/abs-2208-08954,DBLP:journals/corr/abs-2108-10604}, indicating that using \textit{max-pooling} or \textit{average-pooling} over label names directly is a sub-optimal approach. We demonstrate that using \textit{input-prompt}, Mask Matching outperforms Semantic Matching, indicating our method’s better ability to extract semantic information from label names.

\subsection{Low-resource Setting}\label{sec:low_resource}

We also want to explore whether Mask Matching still performs well when the training data is insufficient. The results are shown in Table \ref{tabel:low}, where we only report the results of Prompt-tuning and Mask Matching since Fine-tuning and Semantic Matching can not achieve desirable performances in most low-resource scenarios. From Table \ref{tabel:low} we can see that Mask Matching outperforms Prompt-tuning on 12 of 14 datasets, and the improvements are significant in most cases, especially in entity typing and relation classification. Label names in the above tasks comprise rich semantic information; therefore, with the aid of the \textit{label-prompt}, Mask Matching can take advantage of that information and achieve better results. Improvements in original training sizes that are extensive, such as QNLI, SNLI, and QQP, are considerably minor. In Conclusion, Mask Matching is indeed useful in cases where the label number is significant and label names contain rich semantic information, resulting in a substantial improvement in performance.

\subsection{The Sensitivity to Different Label-Prompts}\label{sec:diff_prompt}

\begin{figure}[h]
	\centering
	\begin{subfigure}{0.49\linewidth}
		\centering
		\includegraphics[width=0.9\linewidth]{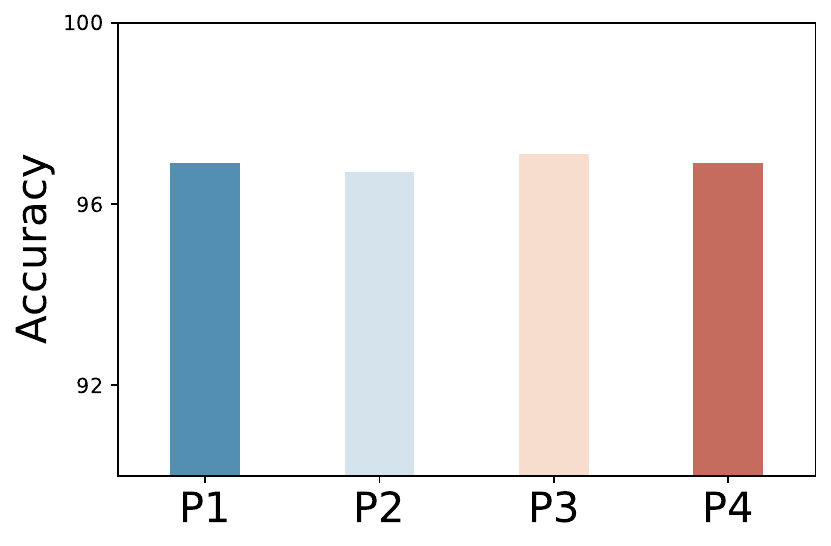}
		\caption{R52}
	\end{subfigure}
	\centering
	\begin{subfigure}{0.49\linewidth}
		\centering
		\includegraphics[width=0.9\linewidth]{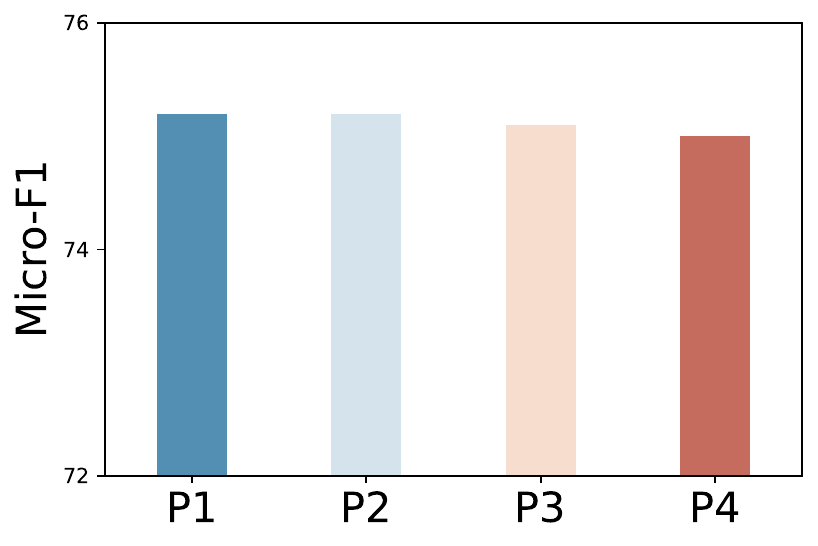}
		\caption{TACRED}
	\end{subfigure}

	\centering
	\begin{subfigure}{0.49\linewidth}
		\centering
		\includegraphics[width=0.9\linewidth]{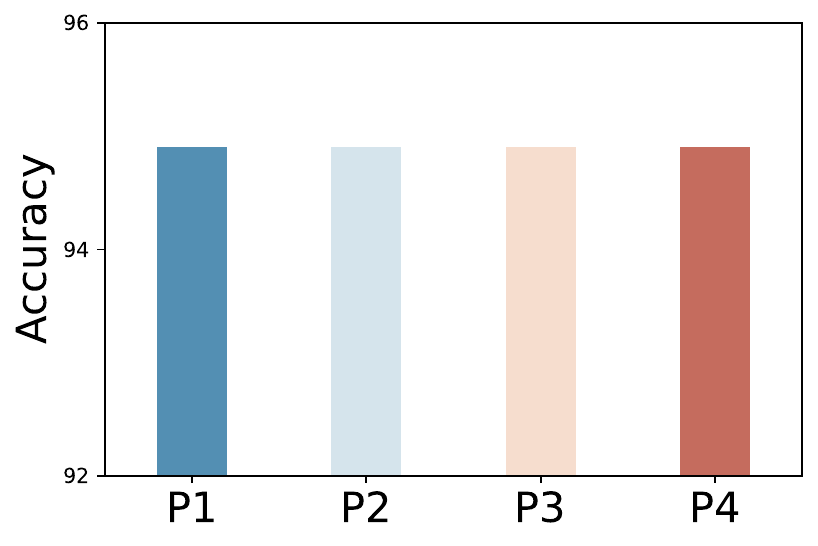}
		\caption{QNLI}
	\end{subfigure}
	\centering
	\begin{subfigure}{0.49\linewidth}
		\centering
		\includegraphics[width=0.9\linewidth]{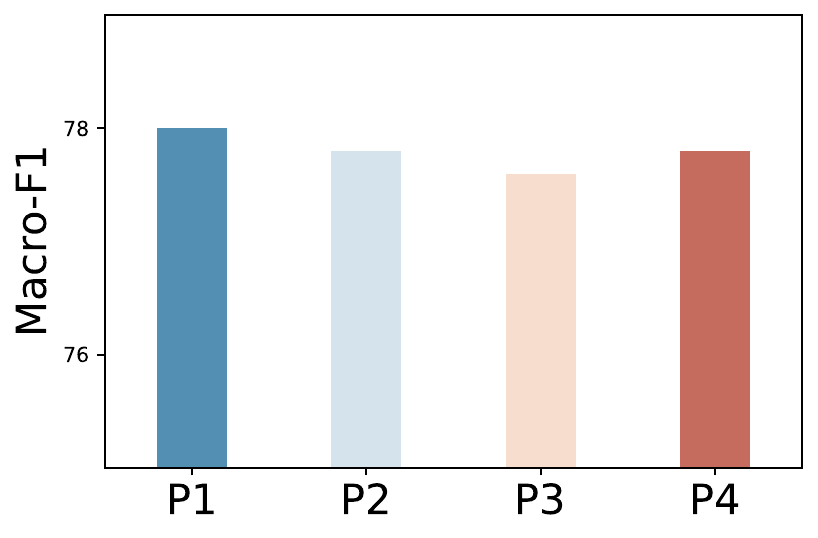}
		\caption{VAST}
	\end{subfigure}
	\caption{The sensitivity to different label-prompts. Mask Matching is insensitive to the template used.}
	\label{pic:diff_prompt}
\end{figure}

We also investigate the effectiveness of different templates used in the \textit{label-prompt}. The default template is \textit{\textbf{P1} = [Label Name] is [MASK]}, as we discussed in \S \ref{sec:mask_match}. We explore using three additional templates: \textit{\textbf{P2} = The meaning of [Label Name] is [MASK]}, \textit{\textbf{P3} = Label Name] means [MASK]}, \textit{\textbf{P4} = [Label Name] is similar to [MASK]} for comparison purposes. We conduct experiments on both \textit{single-input} and \textit{paired-input} datasets, and the results can be seen in Figure \ref{pic:diff_prompt}. The outcomes indicate that all prompt templates yield similar performances, indicating that different prompt templates have little effect on the results. Furthermore, Mask Matching is immune to various templates in the \textit{label-prompt}. Given that \textbf{P1} delivers more stable results than the others, we select this as the default prompt template.

\subsection{Effects of Enriching Label Names}\label{sec:labe_aug}

\begin{table}[h]
\caption{The experimental results with augmented information on two information extraction tasks. The results show that enriching label names benefits Mask Matching.}
\setlength{\tabcolsep}{4mm}
\centering
\label{table:augmentation}
\begin{tabular}{c|c|c}
\hline
                                 & \textbf{FEW-NERD}   & \textbf{TACRED}     \\ \hline
\textbf{Mask Matching}           & 85.2                & 75.2                \\
+ Augmentation & \textbf{85.6(+0.4)} & \textbf{75.4(+0.2)} \\ \hline
\end{tabular}
\end{table}

Previous results in \S \ref{sec:full} and \S \ref{sec:low_resource} suggest that Mask Matching is efficient in leveraging the semantic information contained in label names. Motivated by \citet{DBLP:journals/corr/abs-2212-14266}, we explore the possibility of enhancing our approach by adding more related information to label names. To test this, we use FEW-NERD and TACRED datasets, which belong to information extraction tasks and have label names containing important information for prediction. We manually selected two related words for each label as augmentation information, and combined them with the original label name to form the augmented label. Results displayed in Table \ref{table:augmentation} indicate that incorporating additional information into label names contributes to performance improvement, further demonstrating the utility of label names in the Mask Matching model.

\section{Discussions}\label{sec:discussion}

Despite the promising results achieved by Mask Matching, there is still ample opportunity to enhance our method further. To this end, we discuss multiple research directions in this section, with the goal of encouraging readers to consider the broader use of \textit{label-prompt}.

\textbf{Designing a new framework for paired-input tasks.} As we observed in \S \ref{sec:full} and \S \ref{sec:low_resource}, Mask Matching does not gain remarkable improvements when dealing with sentence-pair tasks, such as natural language processing and paraphrase. Most prompt-tuning methods (including Mask Matching) directly concatenate two sentences and use a mask token for prediction, which may not be the best choice. One potential solution is to represent two sentences using two mask tokens with joint encoding, and then design a new interactive module or training strategy to get the final prediction via two mask representations.

\textbf{Automatically extending the label names.} Although \S \ref{sec:labe_aug} highlights the potential performance improvements resulting from the inclusion of additional relevant information in label names, manually identifying synonyms is time-consuming and challenging without domain knowledge. Automating the extension of label names is, therefore, a promising research direction that could make Mask Matching more robust and generalizable. Furthermore, since the use of Mask Matching requires instances of named labels, extracting important details from input texts and generating appropriate label names automatically is an interesting research area.

\textbf{Exploring collaborative training using multiple mask tokens.} Some works \cite{DBLP:journals/corr/abs-2204-13413,DBLP:conf/acl/ParkJKKN22} utilize multiple mask tokens for downstream tasks and achieve favorable performances. In this research, we only use one mask token in the \textit{input-prompt} and \textit{label-prompt}. We believe using multiple mask tokens and designing a proper training strategy could achieve better performance.


\section{Conclusion} 

We have presented Mask Matching, a paradigm that matches a mask representation generated from a \textit{input-prompt} with another from a \textit{label-prompt}, to uniformly make predictions for a wide range of natural language understanding tasks. Experimental results on a full training setting show that Mask Matching significantly outperforms its fine-tuning and prompt-tuning counterparts, and obtains on-pair or better results than many state-of-the-art methods. Evaluations in the low-resource scenario and several ablation studies further verify the effectiveness of our method. Our research provides a new perspective to utilize the label semantic information, and we hope Mask Matching could inspire more exploration of the prompting methodology on the label side.

\bibliography{aaai24}

\begin{thebibliography}{63}
\providecommand{\natexlab}[1]{#1}

\bibitem[{Allaway and McKeown(2020)}]{allaway2020zero}
Allaway, E.; and McKeown, K. 2020.
\newblock Zero-shot stance detection: A dataset and model using generalized topic representations.
\newblock \emph{CoRR}.

\bibitem[{Alt, Gabryszak, and Hennig(2020)}]{DBLP:conf/acl/AltGH20a}
Alt, C.; Gabryszak, A.; and Hennig, L. 2020.
\newblock {TACRED} Revisited: {A} Thorough Evaluation of the {TACRED} Relation Extraction Task.
\newblock In Jurafsky, D.; Chai, J.; Schluter, N.; and Tetreault, J.~R., eds., \emph{{ACL}}.

\bibitem[{Bajaj et~al.(2022)Bajaj, Xiong, Ke, Liu, He, Tiwary, Liu, Bennett, Song, and Gao}]{DBLP:journals/corr/abs-2204-06644}
Bajaj, P.; Xiong, C.; Ke, G.; Liu, X.; He, D.; Tiwary, S.; Liu, T.; Bennett, P.; Song, X.; and Gao, J. 2022.
\newblock {METRO:} Efficient Denoising Pretraining of Large Scale Autoencoding Language Models with Model Generated Signals.
\newblock \emph{CoRR}.

\bibitem[{Bowman et~al.(2015)Bowman, Angeli, Potts, and Manning}]{DBLP:conf/emnlp/BowmanAPM15}
Bowman, S.~R.; Angeli, G.; Potts, C.; and Manning, C.~D. 2015.
\newblock A large annotated corpus for learning natural language inference.
\newblock In M{\`{a}}rquez, L.; Callison{-}Burch, C.; Su, J.; Pighin, D.; and Marton, Y., eds., \emph{{EMNLP}}.

\bibitem[{Brown et~al.(2020)Brown, Mann, Ryder, Subbiah, Kaplan, Dhariwal, Neelakantan, Shyam, Sastry, Askell, Agarwal, Herbert{-}Voss, Krueger, Henighan, Child, Ramesh, Ziegler, Wu, Winter, Hesse, Chen, Sigler, Litwin, Gray, Chess, Clark, Berner, McCandlish, Radford, Sutskever, and Amodei}]{DBLP:conf/nips/BrownMRSKDNSSAA20}
Brown, T.~B.; Mann, B.; Ryder, N.; Subbiah, M.; Kaplan, J.; Dhariwal, P.; Neelakantan, A.; Shyam, P.; Sastry, G.; Askell, A.; Agarwal, S.; Herbert{-}Voss, A.; Krueger, G.; Henighan, T.; Child, R.; Ramesh, A.; Ziegler, D.~M.; Wu, J.; Winter, C.; Hesse, C.; Chen, M.; Sigler, E.; Litwin, M.; Gray, S.; Chess, B.; Clark, J.; Berner, C.; McCandlish, S.; Radford, A.; Sutskever, I.; and Amodei, D. 2020.
\newblock Language Models are Few-Shot Learners.
\newblock In Larochelle, H.; Ranzato, M.; Hadsell, R.; Balcan, M.; and Lin, H., eds., \emph{{NeurIPS}}.

\bibitem[{Chen et~al.(2022)Chen, Zhang, Xie, Deng, Yao, Tan, Huang, Si, and Chen}]{DBLP:conf/www/ChenZXDYTHSC22}
Chen, X.; Zhang, N.; Xie, X.; Deng, S.; Yao, Y.; Tan, C.; Huang, F.; Si, L.; and Chen, H. 2022.
\newblock KnowPrompt: Knowledge-aware Prompt-tuning with Synergistic Optimization for Relation Extraction.
\newblock In Laforest, F.; Troncy, R.; Simperl, E.; Agarwal, D.; Gionis, A.; Herman, I.; and M{\'{e}}dini, L., eds., \emph{{WWW}}.

\bibitem[{Dawkins(2021)}]{DBLP:conf/acl/Dawkins21}
Dawkins, H. 2021.
\newblock Marked Attribute Bias in Natural Language Inference.
\newblock In Zong, C.; Xia, F.; Li, W.; and Navigli, R., eds., \emph{{Findings of ACL/IJCNLP}}.

\bibitem[{del Arco, Valdivia, and Klinger(2022)}]{DBLP:conf/coling/ArcoVK22}
del Arco, F. M.~P.; Valdivia, M. T.~M.; and Klinger, R. 2022.
\newblock Natural Language Inference Prompts for Zero-shot Emotion Classification in Text across Corpora.
\newblock In \emph{{COLING}}.

\bibitem[{Devlin et~al.(2019)Devlin, Chang, Lee, and Toutanova}]{DBLP:conf/naacl/DevlinCLT19}
Devlin, J.; Chang, M.; Lee, K.; and Toutanova, K. 2019.
\newblock {BERT:} Pre-training of Deep Bidirectional Transformers for Language Understanding.
\newblock In \emph{{NAACL-HLT}}.

\bibitem[{Ding et~al.(2021{\natexlab{a}})Ding, Chen, Han, Xu, Xie, Zheng, Liu, Li, and Kim}]{DBLP:journals/corr/abs-2108-10604}
Ding, N.; Chen, Y.; Han, X.; Xu, G.; Xie, P.; Zheng, H.; Liu, Z.; Li, J.; and Kim, H. 2021{\natexlab{a}}.
\newblock Prompt-Learning for Fine-Grained Entity Typing.
\newblock \emph{CoRR}.

\bibitem[{Ding et~al.(2021{\natexlab{b}})Ding, Xu, Chen, Wang, Han, Xie, Zheng, and Liu}]{DBLP:conf/acl/DingXCWHXZL20}
Ding, N.; Xu, G.; Chen, Y.; Wang, X.; Han, X.; Xie, P.; Zheng, H.; and Liu, Z. 2021{\natexlab{b}}.
\newblock Few-NERD: {A} Few-shot Named Entity Recognition Dataset.
\newblock In Zong, C.; Xia, F.; Li, W.; and Navigli, R., eds., \emph{{ACL/IJCNLP}}.

\bibitem[{Ding et~al.(2021{\natexlab{c}})Ding, Shang, Wang, Sun, Tian, Wu, and Wang}]{DBLP:conf/acl/DingSWSTW020}
Ding, S.; Shang, J.; Wang, S.; Sun, Y.; Tian, H.; Wu, H.; and Wang, H. 2021{\natexlab{c}}.
\newblock ERNIE-Doc: {A} Retrospective Long-Document Modeling Transformer.
\newblock In Zong, C.; Xia, F.; Li, W.; and Navigli, R., eds., \emph{{ACL/IJCNLP}}.

\bibitem[{Gao, Fisch, and Chen(2021)}]{DBLP:conf/acl/GaoFC20}
Gao, T.; Fisch, A.; and Chen, D. 2021.
\newblock Making Pre-trained Language Models Better Few-shot Learners.
\newblock In \emph{{ACL/IJCNLP}}.

\bibitem[{Goyal et~al.(2017)Goyal, Doll{\'{a}}r, Girshick, Noordhuis, Wesolowski, Kyrola, Tulloch, Jia, and He}]{DBLP:journals/corr/GoyalDGNWKTJH17}
Goyal, P.; Doll{\'{a}}r, P.; Girshick, R.~B.; Noordhuis, P.; Wesolowski, L.; Kyrola, A.; Tulloch, A.; Jia, Y.; and He, K. 2017.
\newblock Accurate, Large Minibatch {SGD:} Training ImageNet in 1 Hour.
\newblock \emph{CoRR}.

\bibitem[{Han et~al.(2021)Han, Zhao, Ding, Liu, and Sun}]{DBLP:journals/corr/abs-2105-11259}
Han, X.; Zhao, W.; Ding, N.; Liu, Z.; and Sun, M. 2021.
\newblock {PTR:} Prompt Tuning with Rules for Text Classification.
\newblock \emph{CoRR}.

\bibitem[{He, Gao, and Chen(2021)}]{DBLP:journals/corr/abs-2111-09543}
He, P.; Gao, J.; and Chen, W. 2021.
\newblock DeBERTaV3: Improving DeBERTa using ELECTRA-Style Pre-Training with Gradient-Disentangled Embedding Sharing.
\newblock \emph{CoRR}.

\bibitem[{Huang, Meng, and Han(2022)}]{DBLP:conf/kdd/Huang0022}
Huang, J.; Meng, Y.; and Han, J. 2022.
\newblock Few-Shot Fine-Grained Entity Typing with Automatic Label Interpretation and Instance Generation.
\newblock In Zhang, A.; and Rangwala, H., eds., \emph{{SIGKDD}}.

\bibitem[{Huang et~al.(2022)Huang, Li, Xu, and Chen}]{DBLP:conf/naacl/HuangLXC22}
Huang, J.~Y.; Li, B.; Xu, J.; and Chen, M. 2022.
\newblock Unified Semantic Typing with Meaningful Label Inference.
\newblock In \emph{{NAACL}}.

\bibitem[{Jiang et~al.(2022{\natexlab{a}})Jiang, Huang, Zhang, Wang, Zhuang, Wei, Huang, Zhang, and Zhang}]{DBLP:journals/corr/abs-2201-04337}
Jiang, T.; Huang, S.; Zhang, Z.; Wang, D.; Zhuang, F.; Wei, F.; Huang, H.; Zhang, L.; and Zhang, Q. 2022{\natexlab{a}}.
\newblock PromptBERT: Improving {BERT} Sentence Embeddings with Prompts.
\newblock \emph{CoRR}.

\bibitem[{Jiang et~al.(2022{\natexlab{b}})Jiang, Gao, Shen, and Cheng}]{DBLP:conf/sigir/JiangGSC22}
Jiang, Y.; Gao, J.; Shen, H.; and Cheng, X. 2022{\natexlab{b}}.
\newblock Few-Shot Stance Detection via Target-Aware Prompt Distillation.
\newblock In \emph{{SIGIR}}.

\bibitem[{Lee et~al.(2022)Lee, Kadakia, Tan, Agarwal, Feng, Shibuya, Mitani, Sekiya, Pujara, and Ren}]{DBLP:conf/acl/LeeKTAF0MSPR22}
Lee, D.; Kadakia, A.; Tan, K.; Agarwal, M.; Feng, X.; Shibuya, T.; Mitani, R.; Sekiya, T.; Pujara, J.; and Ren, X. 2022.
\newblock Good Examples Make {A} Faster Learner: Simple Demonstration-based Learning for Low-resource {NER}.
\newblock In Muresan, S.; Nakov, P.; and Villavicencio, A., eds., \emph{{ACL}}.

\bibitem[{Lester, Al{-}Rfou, and Constant(2021)}]{DBLP:conf/emnlp/LesterAC21}
Lester, B.; Al{-}Rfou, R.; and Constant, N. 2021.
\newblock The Power of Scale for Parameter-Efficient Prompt Tuning.
\newblock In \emph{{EMNLP}}.

\bibitem[{Li et~al.(2020)Li, Ye, Sheng, Xie, Xi, and Zhang}]{DBLP:conf/coling/LiYSXXZ20}
Li, B.; Ye, W.; Sheng, Z.; Xie, R.; Xi, X.; and Zhang, S. 2020.
\newblock Graph Enhanced Dual Attention Network for Document-Level Relation Extraction.
\newblock In Scott, D.; Bel, N.; and Zong, C., eds., \emph{{COLING}}.

\bibitem[{Li et~al.(2023)Li, Ye, Zhang, and Zhang}]{DBLP:conf/aaai/Li0ZZ23}
Li, B.; Ye, W.; Zhang, J.; and Zhang, S. 2023.
\newblock Reviewing Labels: Label Graph Network with Top-k Prediction Set for Relation Extraction.
\newblock In Williams, B.; Chen, Y.; and Neville, J., eds., \emph{{AAAI}}.

\bibitem[{Li et~al.(2022)Li, Yu, Ye, Zhang, and Zhang}]{DBLP:journals/corr/abs-2212-14266}
Li, B.; Yu, D.; Ye, W.; Zhang, J.; and Zhang, S. 2022.
\newblock Sequence Generation with Label Augmentation for Relation Extraction.
\newblock \emph{CoRR}.

\bibitem[{Liang et~al.(2022)Liang, Zhu, Li, Yang, Gui, He, and Xu}]{DBLP:conf/acl/LiangZL000X22}
Liang, B.; Zhu, Q.; Li, X.; Yang, M.; Gui, L.; He, Y.; and Xu, R. 2022.
\newblock JointCL: {A} Joint Contrastive Learning Framework for Zero-Shot Stance Detection.
\newblock In \emph{{ACL}}.

\bibitem[{Lin et~al.(2021)Lin, Meng, Sun, Han, Kuang, Li, and Wu}]{DBLP:journals/corr/abs-2105-05727}
Lin, Y.; Meng, Y.; Sun, X.; Han, Q.; Kuang, K.; Li, J.; and Wu, F. 2021.
\newblock BertGCN: Transductive Text Classification by Combining {GCN} and {BERT}.
\newblock \emph{CoRR}.

\bibitem[{Liu, Chen, and Xu(2022)}]{DBLP:conf/ijcai/LiuCX22}
Liu, J.; Chen, Y.; and Xu, J. 2022.
\newblock Low-Resource {NER} by Data Augmentation With Prompting.
\newblock In \emph{{IJCAI}}.

\bibitem[{Liu et~al.(2021{\natexlab{a}})Liu, Yuan, Fu, Jiang, Hayashi, and Neubig}]{DBLP:journals/corr/abs-2107-13586}
Liu, P.; Yuan, W.; Fu, J.; Jiang, Z.; Hayashi, H.; and Neubig, G. 2021{\natexlab{a}}.
\newblock Pre-train, Prompt, and Predict: {A} Systematic Survey of Prompting Methods in Natural Language Processing.
\newblock \emph{CoRR}.

\bibitem[{Liu et~al.(2021{\natexlab{b}})Liu, Liu, Collier, Korhonen, and Vulic}]{DBLP:conf/conll/Liu0CKV21}
Liu, Q.; Liu, F.; Collier, N.; Korhonen, A.; and Vulic, I. 2021{\natexlab{b}}.
\newblock MirrorWiC: On Eliciting Word-in-Context Representations from Pretrained Language Models.
\newblock In Bisazza, A.; and Abend, O., eds., \emph{{CoNLL}}.

\bibitem[{Liu et~al.(2021{\natexlab{c}})Liu, Lin, Tan, and Wang}]{DBLP:conf/acl/LiuLTW21}
Liu, R.; Lin, Z.; Tan, Y.; and Wang, W. 2021{\natexlab{c}}.
\newblock Enhancing Zero-shot and Few-shot Stance Detection with Commonsense Knowledge Graph.
\newblock In \emph{{ACL/IJCNLP}}.

\bibitem[{Liu et~al.(2021{\natexlab{d}})Liu, Zheng, Du, Ding, Qian, Yang, and Tang}]{DBLP:journals/corr/abs-2103-10385}
Liu, X.; Zheng, Y.; Du, Z.; Ding, M.; Qian, Y.; Yang, Z.; and Tang, J. 2021{\natexlab{d}}.
\newblock {GPT} Understands, Too.
\newblock \emph{CoRR}.

\bibitem[{Liu et~al.(2019)Liu, Ott, Goyal, Du, Joshi, Chen, Levy, Lewis, Zettlemoyer, and Stoyanov}]{DBLP:journals/corr/abs-1907-11692}
Liu, Y.; Ott, M.; Goyal, N.; Du, J.; Joshi, M.; Chen, D.; Levy, O.; Lewis, M.; Zettlemoyer, L.; and Stoyanov, V. 2019.
\newblock RoBERTa: {A} Robustly Optimized {BERT} Pretraining Approach.
\newblock \emph{CoRR}.

\bibitem[{Loshchilov and Hutter(2019)}]{DBLP:conf/iclr/LoshchilovH19}
Loshchilov, I.; and Hutter, F. 2019.
\newblock Decoupled Weight Decay Regularization.
\newblock In \emph{{ICLR}}.

\bibitem[{Lu et~al.(2022)Lu, Liu, Dai, Xiao, Lin, Han, Sun, and Wu}]{DBLP:conf/acl/0001LDXLHSW22}
Lu, Y.; Liu, Q.; Dai, D.; Xiao, X.; Lin, H.; Han, X.; Sun, L.; and Wu, H. 2022.
\newblock Unified Structure Generation for Universal Information Extraction.
\newblock In \emph{{ACL}}.

\bibitem[{Maas et~al.(2011)Maas, Daly, Pham, Huang, Ng, and Potts}]{DBLP:conf/acl/MaasDPHNP11}
Maas, A.~L.; Daly, R.~E.; Pham, P.~T.; Huang, D.; Ng, A.~Y.; and Potts, C. 2011.
\newblock Learning Word Vectors for Sentiment Analysis.
\newblock In Lin, D.; Matsumoto, Y.; and Mihalcea, R., eds., \emph{{ACL}}.

\bibitem[{Nighojkar and Licato(2021)}]{DBLP:conf/acl/NighojkarL20}
Nighojkar, A.; and Licato, J. 2021.
\newblock Improving Paraphrase Detection with the Adversarial Paraphrasing Task.
\newblock In Zong, C.; Xia, F.; Li, W.; and Navigli, R., eds., \emph{{ACL/IJCNLP}}.

\bibitem[{Obeidat et~al.(2019)Obeidat, Fern, Shahbazi, and Tadepalli}]{DBLP:conf/naacl/ObeidatFST19}
Obeidat, R.; Fern, X.~Z.; Shahbazi, H.; and Tadepalli, P. 2019.
\newblock Description-Based Zero-shot Fine-Grained Entity Typing.
\newblock In \emph{{NAACL-HLT}}.

\bibitem[{Pang and Lee(2005)}]{DBLP:conf/acl/PangL05}
Pang, B.; and Lee, L. 2005.
\newblock Seeing Stars: Exploiting Class Relationships for Sentiment Categorization with Respect to Rating Scales.
\newblock In Knight, K.; Ng, H.~T.; and Oflazer, K., eds., \emph{{ACL}}.

\bibitem[{Park et~al.(2022)Park, Jeon, Kim, Kang, and Na}]{DBLP:conf/acl/ParkJKKN22}
Park, E.; Jeon, D.~H.; Kim, S.; Kang, I.; and Na, S. 2022.
\newblock {LM-BFF-MS:} Improving Few-Shot Fine-tuning of Language Models based on Multiple Soft Demonstration Memory.
\newblock In \emph{{ACL}}.

\bibitem[{Paszke et~al.(2019)Paszke, Gross, Massa, Lerer, Bradbury, Chanan, Killeen, Lin, Gimelshein, Antiga, Desmaison, K{\"{o}}pf, Yang, DeVito, Raison, Tejani, Chilamkurthy, Steiner, Fang, Bai, and Chintala}]{DBLP:conf/nips/PaszkeGMLBCKLGA19}
Paszke, A.; Gross, S.; Massa, F.; Lerer, A.; Bradbury, J.; Chanan, G.; Killeen, T.; Lin, Z.; Gimelshein, N.; Antiga, L.; Desmaison, A.; K{\"{o}}pf, A.; Yang, E.~Z.; DeVito, Z.; Raison, M.; Tejani, A.; Chilamkurthy, S.; Steiner, B.; Fang, L.; Bai, J.; and Chintala, S. 2019.
\newblock PyTorch: An Imperative Style, High-Performance Deep Learning Library.
\newblock In Wallach, H.~M.; Larochelle, H.; Beygelzimer, A.; d'Alch{\'{e}}{-}Buc, F.; Fox, E.~B.; and Garnett, R., eds., \emph{{NeurIPS}}.

\bibitem[{Pilehvar and Camacho{-}Collados(2019)}]{DBLP:conf/naacl/PilehvarC19}
Pilehvar, M.~T.; and Camacho{-}Collados, J. 2019.
\newblock WiC: the Word-in-Context Dataset for Evaluating Context-Sensitive Meaning Representations.
\newblock In Burstein, J.; Doran, C.; and Solorio, T., eds., \emph{{NAACL-HLT}}.

\bibitem[{Rajpurkar et~al.(2016)Rajpurkar, Zhang, Lopyrev, and Liang}]{DBLP:conf/emnlp/RajpurkarZLL16}
Rajpurkar, P.; Zhang, J.; Lopyrev, K.; and Liang, P. 2016.
\newblock SQuAD: 100, 000+ Questions for Machine Comprehension of Text.
\newblock In Su, J.; Carreras, X.; and Duh, K., eds., \emph{{EMNLP}}.

\bibitem[{Reimers and Gurevych(2019)}]{DBLP:conf/emnlp/ReimersG19}
Reimers, N.; and Gurevych, I. 2019.
\newblock Sentence-BERT: Sentence Embeddings using Siamese BERT-Networks.
\newblock In \emph{{EMNLP-IJCNLP}}.

\bibitem[{Sainz et~al.(2021)Sainz, de~Lacalle, Labaka, Barrena, and Agirre}]{DBLP:conf/emnlp/SainzLLBA21}
Sainz, O.; de~Lacalle, O.~L.; Labaka, G.; Barrena, A.; and Agirre, E. 2021.
\newblock Label Verbalization and Entailment for Effective Zero and Few-Shot Relation Extraction.
\newblock In \emph{{EMNLP}}.

\bibitem[{Schick and Sch{\"{u}}tze(2021)}]{DBLP:conf/eacl/SchickS21}
Schick, T.; and Sch{\"{u}}tze, H. 2021.
\newblock Exploiting Cloze-Questions for Few-Shot Text Classification and Natural Language Inference.
\newblock In Merlo, P.; Tiedemann, J.; and Tsarfaty, R., eds., \emph{{EACL}}.

\bibitem[{Soares et~al.(2019)Soares, FitzGerald, Ling, and Kwiatkowski}]{DBLP:conf/acl/SoaresFLK19}
Soares, L.~B.; FitzGerald, N.; Ling, J.; and Kwiatkowski, T. 2019.
\newblock Matching the Blanks: Distributional Similarity for Relation Learning.
\newblock In \emph{{ACL}}.

\bibitem[{Tabasi, Rezaee, and Pilehvar(2022)}]{DBLP:conf/acl/TabasiRP22}
Tabasi, M.; Rezaee, K.; and Pilehvar, M.~T. 2022.
\newblock Exploiting Language Model Prompts Using Similarity Measures: {A} Case Study on the Word-in-Context Task.
\newblock In Muresan, S.; Nakov, P.; and Villavicencio, A., eds., \emph{{ACL}}.

\bibitem[{Tian et~al.(2021)Tian, Chen, Song, and Wan}]{DBLP:conf/acl/TianCSW20}
Tian, Y.; Chen, G.; Song, Y.; and Wan, X. 2021.
\newblock Dependency-driven Relation Extraction with Attentive Graph Convolutional Networks.
\newblock In \emph{ACL/IJCNLP}.

\bibitem[{Wang et~al.(2019{\natexlab{a}})Wang, Pruksachatkun, Nangia, Singh, Michael, Hill, Levy, and Bowman}]{DBLP:conf/nips/WangPNSMHLB19}
Wang, A.; Pruksachatkun, Y.; Nangia, N.; Singh, A.; Michael, J.; Hill, F.; Levy, O.; and Bowman, S.~R. 2019{\natexlab{a}}.
\newblock SuperGLUE: {A} Stickier Benchmark for General-Purpose Language Understanding Systems.
\newblock In Wallach, H.~M.; Larochelle, H.; Beygelzimer, A.; d'Alch{\'{e}}{-}Buc, F.; Fox, E.~B.; and Garnett, R., eds., \emph{{NeurIPS}}.

\bibitem[{Wang et~al.(2019{\natexlab{b}})Wang, Singh, Michael, Hill, Levy, and Bowman}]{DBLP:conf/iclr/WangSMHLB19}
Wang, A.; Singh, A.; Michael, J.; Hill, F.; Levy, O.; and Bowman, S.~R. 2019{\natexlab{b}}.
\newblock {GLUE:} {A} Multi-Task Benchmark and Analysis Platform for Natural Language Understanding.
\newblock In \emph{{ICLR}}. OpenReview.net.

\bibitem[{Wang et~al.(2021)Wang, Fang, Khabsa, Mao, and Ma}]{DBLP:journals/corr/abs-2104-14690}
Wang, S.; Fang, H.; Khabsa, M.; Mao, H.; and Ma, H. 2021.
\newblock Entailment as Few-Shot Learner.
\newblock \emph{CoRR}.

\bibitem[{Wang et~al.(2022{\natexlab{a}})Wang, Xu, Sun, Hu, Tao, Geng, and Jiang}]{DBLP:conf/acl/0003XSHTGJ22}
Wang, Y.; Xu, C.; Sun, Q.; Hu, H.; Tao, C.; Geng, X.; and Jiang, D. 2022{\natexlab{a}}.
\newblock PromDA: Prompt-based Data Augmentation for Low-Resource {NLU} Tasks.
\newblock In \emph{{ACL}}.

\bibitem[{Wang et~al.(2022{\natexlab{b}})Wang, Wang, Liu, Cao, Sui, and Wang}]{DBLP:journals/corr/abs-2204-13413}
Wang, Z.; Wang, P.; Liu, T.; Cao, Y.; Sui, Z.; and Wang, H. 2022{\natexlab{b}}.
\newblock {HPT:} Hierarchy-aware Prompt Tuning for Hierarchical Text Classification.
\newblock \emph{CoRR}.

\bibitem[{Weischedel and Brunstein(2005)}]{weischedel2005bbn}
Weischedel, R.; and Brunstein, A. 2005.
\newblock BBN pronoun coreference and entity type corpus.
\newblock \emph{Linguistic Data Consortium, Philadelphia}, 112.

\bibitem[{Wu et~al.(2022)Wu, Gao, Lin, Zang, and Hu}]{DBLP:conf/acl/0002GLZH22}
Wu, X.; Gao, C.; Lin, M.; Zang, L.; and Hu, S. 2022.
\newblock Text Smoothing: Enhance Various Data Augmentation Methods on Text Classification Tasks.
\newblock In \emph{{ACL}}.

\bibitem[{Xu, Liu, and Abbasnejad(2022)}]{DBLP:conf/naacl/XuLA22}
Xu, H.; Liu, L.; and Abbasnejad, E. 2022.
\newblock Progressive Class Semantic Matching for Semi-supervised Text Classification.
\newblock In \emph{{NAACL}}.

\bibitem[{Xu, Callison{-}Burch, and Dolan(2015)}]{DBLP:conf/semeval/XuCD15}
Xu, W.; Callison{-}Burch, C.; and Dolan, B. 2015.
\newblock SemEval-2015 Task 1: Paraphrase and Semantic Similarity in Twitter {(PIT)}.
\newblock In Cer, D.~M.; Jurgens, D.; Nakov, P.; and Zesch, T., eds., \emph{{SemEval@NAACL-HLT}}.

\bibitem[{Zhang et~al.(2022)Zhang, Zhang, Zhang, Zhao, Liu, Wu, and Chen}]{DBLP:conf/acl/ZhangZZZL0C22}
Zhang, K.; Zhang, K.; Zhang, M.; Zhao, H.; Liu, Q.; Wu, W.; and Chen, E. 2022.
\newblock Incorporating Dynamic Semantics into Pre-Trained Language Model for Aspect-based Sentiment Analysis.
\newblock In \emph{Findings of {ACL}}.

\bibitem[{Zhang et~al.(2017)Zhang, Zhong, Chen, Angeli, and Manning}]{DBLP:conf/emnlp/ZhangZCAM17}
Zhang, Y.; Zhong, V.; Chen, D.; Angeli, G.; and Manning, C.~D. 2017.
\newblock Position-aware Attention and Supervised Data Improve Slot Filling.
\newblock In \emph{{EMNLP}}.

\bibitem[{Zhao, Ma, and Lei(2022)}]{DBLP:journals/corr/abs-2208-08954}
Zhao, Q.; Ma, S.; and Lei, Y. 2022.
\newblock Ered: Enhanced Text Representations with Entities and Descriptions.
\newblock \emph{CoRR}.

\bibitem[{Zhong et~al.(2022)Zhong, Gao, Ding, Liu, Zhou, Wang, Yin, and Duan}]{DBLP:journals/corr/abs-2208-03229}
Zhong, W.; Gao, Y.; Ding, N.; Liu, Z.; Zhou, M.; Wang, J.; Yin, J.; and Duan, N. 2022.
\newblock Improving Task Generalization via Unified Schema Prompt.
\newblock \emph{CoRR}.

\bibitem[{Zhou et~al.(2022)Zhou, Ethayarajh, Card, and Jurafsky}]{DBLP:conf/acl/ZhouECJ22}
Zhou, K.; Ethayarajh, K.; Card, D.; and Jurafsky, D. 2022.
\newblock Problems with Cosine as a Measure of Embedding Similarity for High Frequency Words.
\newblock In \emph{{ACL}}.

\end{thebibliography}

\end{document}